%% file: acl_latex.tex
\pdfoutput=1

\documentclass[11pt]{article}

\usepackage[preprint]{acl}

\usepackage{times}
\usepackage{latexsym}
\usepackage{graphicx} 
\usepackage{amsmath}
\usepackage{algorithm}
\usepackage{booktabs}
\usepackage{multirow}
\usepackage{algorithmic}
\usepackage{natbib}  
\usepackage{caption} 
\usepackage[T1]{fontenc}

\usepackage[utf8]{inputenc}

\usepackage{microtype}

\usepackage{inconsolata}

\usepackage{graphicx}
\usepackage{algorithm}
\usepackage{algorithmic}
\usepackage{array}

\usepackage{newfloat}
\usepackage{listings}
\DeclareCaptionStyle{ruled}{labelfont=normalfont,labelsep=colon,strut=off} 
\floatstyle{ruled}
\newfloat{listing}{tb}{lst}{}
\floatname{listing}{Listing}
\usepackage{times}  
\usepackage{helvet}  
\usepackage{courier}  

\title{Predictable Emergent Abilities of LLMs: Proxy Tasks Are All You Need}

\author{
 \textbf{Bo-Wen Zhang\textsuperscript{1,}}\footnote{\thanks{Correspondence: \href{mailto:bowenzhang@baai.ac.cn}{bowenzhang@baai.ac.cn}}},
 \textbf{Yan Yan\textsuperscript{2}},
 \textbf{Boxiang Yang\textsuperscript{2}},
 \textbf{Yifei Xue\textsuperscript{1,3}},
 \textbf{Guang Liu\textsuperscript{1,2}},
\\
 \textsuperscript{1}Beijing Academy of Artificial Intelligence (BAAI),\\
 \textsuperscript{2}China University of Mining and Technology Beijing,
 \textsuperscript{3}Peking University
}

\begin{document}
\maketitle
\begin{abstract}
While scaling laws optimize training configurations for large language models (LLMs) through experiments on smaller or early-stage models, they fail to predict emergent abilities due to the absence of such capabilities in these models. To address this, we propose a method that predicts emergent abilities by leveraging proxy tasks. We begin by establishing relevance metrics between the target task and candidate tasks based on performance differences across multiple models. These candidate tasks are then validated for robustness with small model ensembles, leading to the selection of the most appropriate proxy tasks. The predicted performance on the target task is then derived by integrating the evaluation results of these proxies. In a case study on tool utilization capabilities, our method demonstrated a strong correlation between predicted and actual performance, confirming its effectiveness.
\end{abstract}

\input{latex/introduction}

\input{latex/relatedwork}

\input{latex/methods}

\input{latex/experiments}

\input{latex/conclusion}

\bibliography{latex/acl_latex}




\end{document}

%% file: latex/introduction.tex
\section{Introduction}

Large Language Models (LLMs) demonstrate exceptional capabilities across a range of tasks, yet their success is contingent on substantial model sizes and extensive datasets, necessitating significant computational resources. The outcomes of these models are highly sensitive to variations in configurations, including model-specific tuning, dataset composition, and hyperparameters. Consequently, accurately predicting a model's future performance under specific settings becomes essential. Although scaling laws provide insight into the potential of certain abilities \cite{henighan2020scaling}, they are inadequate to forecast emergent capabilities, which do not manifest themselves in smaller or early-stage models \cite{wei2022emergent}.

Tool utilization stands as a quintessential example of emergent abilities, requiring the integration of complex skills such as instruction following, planning, reasoning, retrieval, comprehension, and iterative review. Accurate prediction of an LLM’s capacity for tool use is critical, as it directly impacts the model's effectiveness in executing sophisticated tasks. However, the evaluation of this capability is challenging due to the above-mentioned limitations in early-stage models, where such advanced abilities are not yet developed, and the models’ sensitivity to training configurations adds further complexity.

To address the challenges in predicting emergent abilities like tool utilization, we propose a novel approach that leverages proxy tasks for early-stage evaluation. Our method begins by establishing relevance metrics between the target task and a pool of candidate tasks. These metrics are derived from performance differences across multiple models, allowing us to identify tasks that share similar underlying requirements with the target task. By focusing on these closely related tasks, we aim to provide a more accurate reflection of the capabilities needed for the emergent ability in question.

Once candidate tasks are identified, they undergo robustness validation using small model ensembles. This step is critical because early stage models often exhibit high sensitivity to noise and other perturbations in the training process. By employing small model ensembles, we can systematically assess the stability and reliability of these tasks under various configurations. Tasks that demonstrate consistent performance across different models and settings are selected as the most suitable proxies, ensuring that the early evaluation is both stable and predictive.

Finally, we synthesize the evaluation results from the selected proxy tasks to predict the performance of the target task. This synthesis involves normalizing and integrating the results to produce a cohesive assessment. As demonstrated in our case study on tool-use capabilities, this approach enables us to predict the final performance of LLMs on complex tasks with a high degree of accuracy. The strong correlation observed between early proxy task evaluations and later-stage performance validates the effectiveness of our method, providing a robust framework for optimizing LLM training configurations in the early stages.

The primary innovation of this work lies in the development of a novel two-stage approach for predicting emergent abilities in LLMs during early-stage training. By introducing a method to construct and validate proxy tasks based on relevance and robustness metrics, we shift the focus of complex task evaluation from later stages to the early phases of model development. This paradigm not only enables more accurate predictions of complex abilities but also provides actionable insights for optimizing training configurations. Our approach offers a significant advancement in the efficient and reliable assessment of LLMs, addressing the limitations of traditional evaluation methods and paving the way for more effective utilization of large-scale models in real-world applications.

%% file: latex/relatedwork.tex
\section{Related Work}

The scaling law, established by \cite{henighan2020scaling}, has significantly influenced the trajectory of the development of large-scale artificial intelligence models. This law articulates the relationship between model size, dataset scale, and computational resources, demonstrating that as these factors increase, the model's performance exhibits a consistent improvement. This principle has become a cornerstone for optimizing and predicting the performance of large language models (LLMs), enabling the creation of high-performing models, such as ChatGPT by OpenAI. However, it is important to acknowledge the diminishing returns on performance gains as the models scale up. Even at the threshold where marginal improvements taper, engineering innovations continue to enhance the quality and semantic richness of generated outputs. OpenAI’s empirical studies affirm that it is feasible to anticipate the performance of larger models based on the outcomes of smaller models, a concept known as ``Predictable Scaling"~\cite{achiam2023gpt}. This predictability can be categorized into parameter scaling~\cite{xie2024doremi} and corpus scaling, contingent on the aspect being scaled.

Beyond predictable scaling, an intriguing phenomenon in large models is the emergence of abilities that are absent in smaller models. These ``emergent abilities" were first highlighted by \cite{wei2022emergent}. Emergent abilities tend to appear unpredictably when a model reaches a critical size, leading to substantial enhancements in complex tasks such as in-context learning~\cite{brown2020language}, chain-of-thought reasoning~\cite{wei2022chain}, and tool-use. Tool-use, as characterized by \cite{ling2023international}, enables LLMs to perform tasks like executing arithmetic operations, making function calls, and web browsing via application programming interfaces (APIs). The dramatic improvements brought about by emergent abilities underscore the importance of their prediction and evaluation. However, predicting the onset of these capabilities remains challenging, as evidenced by the extensive exploration of this topic in the literature~\cite{ganguli2022predictability, suzgun2022challenging, wei2022emergent}. Recent work suggests that this unpredictability may be due to the metrics used~\cite{lu2023emergent, srivastava2022beyond} or the limited number of data points considered~\cite{hu2023predicting, anwar2024foundational}. Further research has delved into the correlation between pretraining loss and downstream performance in language models~\cite{huang2024compression, xia2022training}, contributing to our understanding of scaling and emergent abilities~\cite{gadre2024language, du2024understanding}. A theoretical framework proposed by \cite{arora2023theory} posits that the performance of LLMs on complex tasks can be viewed as a combination of basic competencies.

In the early evaluation stages of large models, diverse methodologies have been employed to ensure comprehensive and precise assessments. Metrics such as perplexity, derived from the concept of entropy in information theory~\cite{shannon1948mathematical}, have been crucial. Lower perplexity values indicate stronger predictive capabilities. The DoReMi study~\cite{xie2024doremi} used perplexity to fine-tune data mixtures, achieving optimal performance with significantly less data prior to training. Similarly, the Falcon series~\cite{almazrouei2023falcon} introduced a suite of downstream task evaluation metrics as an alternative to perplexity for early-stage assessments. This approach, which evaluates models using a set of downstream tasks and averages their scores, provides a practical and interpretable measure of model capabilities, making it particularly valuable during the preliminary stages of model development, despite its inherent subjectivity.

%% file: latex/methods.tex
\section{Methodology}

\subsection{Relevance-Based Metric for Proxy Task Selection}

In proxy task selection, the primary goal is to identify tasks that closely align with the target task. This alignment is determined through relevance, evaluated using performance features derived from various models. We introduce a novel relevance evaluation method that leverages performance data obtained from multiple models on well-established evaluation benchmarks, which serve as the candidate set for proxy tasks. These benchmarks are selected because they not only reflect the fundamental capabilities of models but also have been validated in scaling law studies. Additionally, the availability of extensive model results on these benchmarks across various leaderboards ensures that the data is easy to obtain and standardized, eliminating the need for extra evaluations. By constructing features based on these results, each task's relevance is quantified, enabling the systematic selection of suitable proxy tasks.

Given a set of $n$ evaluation tasks, $\mathcal{T} = \{T_1, T_2, \dots, T_n\}$, and a set of $m$ models, $\mathcal{M} = \{M_1, \dots, M_m\}$, let $s_{ij}$ represent the performance score of the model $M_i$ on task $T_j$ (e.g. accuracy, F1 score). For each task $T_j$, we construct an $m$-dimensional performance vector:

\[
\mathbf{p}_j = [s_{1j}, s_{2j}, \dots, s_{mj}]^T
\]

This vector $\mathbf{p}_j$ captures the performance characteristics of task $T_j$ across multiple models. To ensure consistency and comparability across tasks and models, it is essential to normalize these performance vectors. Normalization can be applied both across tasks (feature normalization) and across models (sample normalization) to account for varying scales and task difficulties.

To achieve feature normalization, we normalize each task's performance scores across models. Let $\mu_j$ and $\sigma_j$ represent the mean and standard deviation of the $j$-th task's performance vector, respectively. The feature-normalized score $p_{ij}^{(f)}$ is:

\[
p_{ij}^{(f)} = \frac{p_{ij} - \mu_j}{\sigma_j}
\]

Next, for sample normalization, which normalizes each model's performance across different tasks, let $\mu_i$ and $\sigma_i$ represent the mean and standard deviation of the $i$-th model's performance across all tasks. The sample-normalized score $p_{ij}^{(s)}$ is:

\[
p_{ij}^{(s)} = \frac{p_{ij}^{(f)} - \mu_i}{\sigma_i}
\]

This two-step normalization process—first by tasks and then by models—results in a refined performance matrix that stabilizes and balances the representation of each task and model. The combined normalization ensures that variations due to differing task difficulties or model scales are minimized, allowing for a more consistent and accurate evaluation of tasks.

Finally, the normalized performance vectors can be aggregated into an $m \times n$ matrix $\mathbf{P}$:

\[
\mathbf{P} = 
\begin{bmatrix}
p_{11}^{(s)} & p_{12}^{(s)} & \dots & p_{1n}^{(s)} \\
p_{21}^{(s)} & p_{22}^{(s)} & \dots & p_{2n}^{(s)} \\
\vdots & \vdots & \ddots & \vdots \\
p_{m1}^{(s)} & p_{m2}^{(s)} & \dots & p_{mn}^{(s)}
\end{bmatrix}
\]

This matrix serves as the foundation for further analysis, allowing a systematic comparison of tasks and models based on their normalized performance characteristics.

\begin{algorithm}
\caption{Sampling Consistency Evaluation}
\footnotesize
\label{alg:sampling_consistency}
\begin{algorithmic}[1]
\REQUIRE Task set \(\mathcal{T}\), model set \(\mathcal{M}\), correlation metric set \(\mathcal{C}\), baseline task \(T_b\), sampling count \(k\), models sampled each time \(n\), top tasks of interest \(t\)
\ENSURE Baseline consistency index \(s\), sampling consistency index \(r\)
\FOR{each \(c_i \in \mathcal{C}\)}
    \STATE Compute the relevance of \(T_b\) with all models in \(\mathcal{M}\) to obtain the baseline ranking \(\mathbf{p}_i^{(0)}\)
    \FOR{\(j = 1 \) to \( k \) }
        \STATE Randomly select \(n\) models from \(\mathcal{M}\) as \(\mathcal{M}_j\)
        \STATE Compute the relevance of \(T_b\) with models in \(\mathcal{M}_j\) to obtain ranking \(\mathbf{p}_i^{(j)}\)
        \STATE Calculate the overlap ratio \(o_i^{(j)}\) of the top \(t\) tasks between \(\mathbf{p}_i^{(0)}\) and \(\mathbf{p}_i^{(j)}\)
    \ENDFOR
    \STATE Compute the average overlap ratio as the baseline consistency index \( s_i = \frac{1}{k} \sum_{j=1}^{k} o_i^{(j)} \)
    \FOR{\( j_1 = 1 \) to \( k-1 \)}
        \FOR{\( j_2 = j_1 + 1 \) to \( k \)}
            \STATE Calculate the overlap ratio \( q_i^{(j_1,j_2)} \) of the top \( t \) tasks between rankings \(\mathbf{p}_i^{(j_1)}\) and \(\mathbf{p}_i^{(j_2)}\)
        \ENDFOR
    \ENDFOR
    \STATE Compute the average overlap ratio between sampled rankings as the sampling consistency index \( r_i = \frac{2}{k(k-1)} \sum_{j_1=1}^{k-1} \sum_{j_2=j_1+1}^{k} q_i^{(j_1,j_2)} \)
\ENDFOR
\RETURN \( \mathbf{s} = [s_1, s_2, \ldots], \mathbf{r} = [r_1, r_2, \ldots] \)
\end{algorithmic}
\end{algorithm}

Common metrics for assessing correlation include the Pearson correlation coefficient, the Spearman rank correlation coefficient, and the Kendall rank correlation coefficient. However, given that the task features are derived from the performance of multiple models, the correlation is heavily influenced by the selection of these models. Consequently, rather than relying on a fixed correlation metric, we employ baseline consistency and sampling consistency to evaluate and compare correlation metrics across a specified set of models. This approach allows us to determine the most suitable method for measuring relevance in this context.

Specifically, as shown in Algorithm \ref{alg:sampling_consistency}, we first select a task, denoted \( T_b \), to serve as the baseline to evaluate the relevance of all other tasks. For each correlation metric, the following steps are executed: 

\begin{itemize}
    \item Compute the correlations across all \( m \) models to obtain the baseline relevance ranking.
    \item Perform \( k \) rounds of random sampling, where each round involves randomly selecting a subset of \( m \) models to recompute the correlations and derive a new relevance ranking.
    \item For each sampled ranking, calculate the overlap ratio of the top \( t \) tasks with the top \( t \) tasks in the baseline ranking. The average of these overlap ratios yields the baseline consistency index \( s_i \) for each metric.
    \item Additionally, compute the overlap ratio between the top \( t \) tasks across pairs of sampled rankings to determine the sampling consistency index \( r_i \).
\end{itemize}

Finally, these indices are employed to assess the robustness of the correlation metrics. If a metric satisfies \( s_i \geq s_j \) and \( r_i > r_j \) for all \( j \neq i \), it is deemed to exhibit stronger robustness.

This flexible method for selecting correlation metrics has several clear advantages. First, using random sampling simulates different model selection scenarios, allowing for a comprehensive evaluation of the stability of the correlation metrics. Second, by comparing the rankings with both the baseline and among the sampled results, this approach provides a well-rounded assessment of consistency. Third, focusing on the top-ranked tasks aligns well with practical applications, as researchers often prioritize the most relevant tasks. Finally, the evaluation indices are simple to calculate and produce results that are intuitive and easy to interpret.

\subsection{Robustness-Based Metric for Proxy Task Selection}

To effectively select a proxy task, it is important not only to consider its relevance to the target task but also to evaluate its robustness to training uncertainties. We propose a task robustness analysis method that uses two groups of small models trained from scratch. The Data Variability Group is trained on different data sources to introduce variations in data distribution, while the Random Noise Group is trained on the same data source but with different initialization seeds, introducing variations due to random noise.

The underlying hypothesis is that if a task exhibits significantly higher variance in the Data Variability Group compared to the Random Noise Group, it suggests that the task is more influenced by data distribution differences rather than random noise, indicating stronger robustness. On the other hand, if a task shows comparable or higher variance in the Random Noise Group, it implies that the task is highly sensitive to random factors, resulting in greater variability and lower robustness.

Given a set of candidate evaluation tasks $\mathcal{T} = \{T_1, T_2, \ldots, T_n\}$, the objective is to assess the robustness of each task $T_i$ to model uncertainty, and then rank and filter tasks based on their robustness scores. We first construct two small model ensembles: the Data Variability Group $\mathcal{D}$ and the Random Noise Group $\mathcal{R}$, defined as:
\[
\mathcal{D} = \{M_{d_1}, M_{d_2}, \ldots, M_{d_k}\}
\]
\[
\mathcal{R} = \{M_{r_1}, M_{r_2}, \ldots, M_{r_k}\}
\]
where $k$ represents the number of models in each group. Models in $\mathcal{D}$ are trained on different data sources $\{D_1, D_2, \ldots, D_k\}$, while models in $\mathcal{R}$ are trained on the same data source but with different initialization seeds $\{s_1, s_2, \ldots, s_k\}$.

For each task $T_i$, performance is evaluated on both groups, producing two sets of performance scores:
\[
p_{d_i} = [p_{d_i1}, p_{d_i2}, \ldots, p_{d_ik}]
\]
\[
p_{r_i} = [p_{r_i1}, p_{r_i2}, \ldots, p_{r_ik}]
\]
where $p_{d_ij}$ denotes the performance of task $T_i$ on the $j$-th model $M_{d_j}$ in $\mathcal{D}$, and $p_{r_ij}$ denotes the performance of task $T_i$ on the $j$-th model $M_{r_j}$ in $\mathcal{R}$. We then calculate the variance of these performance scores for task $T_i$ in both ensembles:
\[
\sigma^2_{d_i} = \frac{1}{k-1} \sum_{j=1}^{k} (p_{d_ij} - \bar{p}_{d_i})^2
\]
\[
\sigma^2_{r_i} = \frac{1}{k-1} \sum_{j=1}^{k} (p_{r_ij} - \bar{p}_{r_i})^2
\]
where $\bar{p}_{d_i}$ and $\bar{p}_{r_i}$ represent the mean performance scores of task $T_i$ in $\mathcal{D}$ and $\mathcal{R}$, respectively. The robustness score $R_i$ for task $T_i$ is then defined as the ratio of these variances:
\[
R_i = \frac{\sigma^2_{d_i}}{\sigma^2_{r_i}}
\]
A task $T_i$ with a higher variance in $\mathcal{D}$ compared to $\mathcal{R}$ is less affected by random noise and more by data distribution, indicating stronger robustness. Such tasks are more reliable for early-stage model evaluation and comparison, providing stable and interpretable evaluation results.

This robustness score highlights tasks that are influenced more by data distribution variations than by random noise, making them ideal as proxy tasks. Tasks with higher robustness scores offer more stable and interpretable results, providing reliable feedback for predicting emergent capabilities during the initial phases of model training.

\subsection{Prediction from Proxy Task Evaluation}

We integrate the correlation analysis between candidate proxy tasks and the target task with the robustness of candidate tasks to training uncertainties. This integration forms a comprehensive evaluation metric for selecting the optimal set of proxy tasks.

Specifically, we first apply thresholds based on task relevance and robustness to preliminarily filter the candidate tasks. We then employ a transformation function that preserves the original ranking of the variance ratios while minimizing the impact of extreme values, preventing outliers from disproportionately influencing the weighting process. Finally, we normalize the weighted scores of all tasks to determine their relative importance. Based on these comprehensive evaluations, the highest-scoring tasks are selected as the final proxy task set. These tasks are not only highly correlated with the target task but also exhibit strong robustness to training uncertainties, ensuring reliable and effective early-stage model evaluation.

Given a set of candidate proxy tasks $\mathcal{T} = \{T_1, T_2, \ldots, T_n\}$ and a target task $T_0$, the objective is to select an optimal subset $\mathcal{S} \subseteq \mathcal{T}$ that enables effective and reliable early-stage model evaluation. For each candidate task $T_i$, we calculate two scores: relevance $C_i$, representing the task's correlation with the target task $T_0$, and robustness $R_i$, reflecting the task's robustness to training uncertainties.

Next, we establish thresholds for relevance $\epsilon_C$ and robustness $\epsilon_R$. Tasks with relevance or robustness scores below these thresholds are discarded, while those above are retained to form a refined proxy task set.

For each task in the refined set $\mathcal{S}$, an evaluation score $S_i$ is computed as the product of its relevance and a transformed robustness score:
\[
S_i = C_i \cdot f(R_i)
\]
where $f(\cdot)$ is a monotonic transformation function, such as the sigmoid function $f(x) = \frac{1}{1 + e^{-kx}}$, which smooths robustness scores. This approach ensures that extreme robustness values do not disproportionately affect the evaluation.

Finally, the relative importance of each task is determined by normalizing the evaluation scores:
\[
W_i = \frac{S_i}{\sum_{j=1}^{n} S_j}
\]

By integrating relevance and robustness scores, this method optimizes the selection of proxy tasks, enabling reliable and effective early-stage model evaluation.

%% file: latex/experiments.tex
\section{Experimental Results and Analysis}
For evaluation, we selected T-eval~\cite{chen2023t} as the benchmark for assessing tool-use capabilities, due to its extensive coverage of various subsets. T-eval has demonstrated comprehensiveness and depth in evaluating large models, with results that accurately reflect their actual performance in this domain. To complement this, we carefully curated a set of 42 diverse evaluation tasks as the candidate proxy task pool (see Table~\ref{tab:tasks}). These tasks encompass a wide range of large language model capabilities, and are predominantly designed in a multiple-choice format with four options, which is particularly advantageous during the early stages of model training. Compared to T-eval, these tasks are more likely to yield valuable evaluation results in the initial training phases, enhancing the timeliness and accuracy of our evaluation system and providing the ability to predict final performance through scaling laws, making them suitable as proxy tasks.

\begin{table}[h]
\centering
\scriptsize
\caption{Candidate Proxy Tasks for Early-Stage Model Evaluation}
\label{tab:tasks}
\begin{tabular}{>{\centering\arraybackslash}m{2.5cm}|>{\centering\arraybackslash}m{5cm}}
\hline
\textbf{Capability} & \textbf{Task List} \\ \hline
Problem-solving     & C-Eval, AGIEval, MMLU, CMMLU, GAOKAO-Bench, ARC-c, ARC-e \\ \hline
Language          & WiC, CHID, AFQMC, WSC, TyDiQA, Flores \\ \hline
Knowledge         & BoolQ, CommonSenseQA, TriviaQA, NaturalQuestions \\ \hline
Comprehension         & C3, RACE(Middle), RACE(High), OpenbookQA, CSL, LCSTS, XSum, EPRSTMT, LAMBADA \\ \hline
Reasoning         & CMNLI, QCNLI, AX-b, AX-g, RTE, COPA, ReCoRD, HellaSwag, PIQA, SIQA, MATH, GSM8K, DROP, HumanEval, MBPP, BBH \\ \hline
\end{tabular}
\end{table}

\subsection{Results for Relevance-based Selection}
This experiment selects 17 pairs of LLMs (a chat model and its corresponding base model, shown in Table~\ref{tab:models}) to construct an m-dimensional performance vector, where \(m = 42\). For the target downstream task (T-eval), evaluations use the chat models' results, as the T-eval evaluation method follows a conversational evaluation paradigm, making base models unsuitable for reliable assessment. The performance on the other 42 candidate tasks is evaluated using both the base models and the chat models, as this better reflects the evaluation methods applicable to early-stage models. The evaluation results for these candidate tasks are obtained from publicly available results via the OpenCompass\cite{buitrago2019open} leaderboard. For T-eval, the results were generated through our own evaluations. The correlation metric selection related to task relevance were conducted based on Algorithm 1, with configurations of \(n=6, 8, 10\), \(k=15, 25, 35\), and \(t=10\).

\begin{table}[h!]
\centering
\scriptsize	
\caption{Models Used for performance vector}
\label{tab:models}
\begin{tabular}{c|c|c}
\hline
\textbf{Chat Model} & \textbf{Base Model} & \textbf{Institution} \\ \hline
Vicuna-7B-v1.5-16k & LLama-7B & LMSYS \& Meta \\ \hline
Vicuna-13B-v1.5-16k & LLama-13B & LMSYS \& Meta \\ \hline
Qwen-7B-Chat & Qwen-7B & Alibaba \\ \hline
Qwen-72B-Chat & Qwen-72B & Alibaba \\ \hline
Qwen-14B-Chat & Qwen-14B & Alibaba \\ \hline
OrionStar-Yi-34B-Chat & Yi-34B & OrionStarAI \& 01.AI \\ \hline
Mixtral-8x7B-Instruct-v0.1 & Mixtral-8x7B-v0.1 & Mistral AI \\ \hline
Mistral-7B-Instruct-v0.2 & Mixtral-7B-v0.1 & Mistral AI \\ \hline
LLAMA-2-7B-Chat & LLama-2-7B & Meta \\ \hline
LLAMA-2-13B-Chat & LLama-2-13B & Meta \\ \hline
LLAMA-2-70B-Chat & LLama-2-70B & Meta \\ \hline
InternLM2-Chat-7B & InternLM2-7B & Shanghai AI Lab \\ \hline
InternLM2-Chat-20B & InternLM2-20B & Shanghai AI Lab \\ \hline
DeepSeek-7B-Chat & DeepSeek-7B-Base & DeepSeek \\ \hline
DeepSeek-67B-Chat & DeepSeek-67B-Base & DeepSeek \\ \hline
Baichuan2-7B-Chat & Baichuan2-7B-Base & Baichuan \\ \hline
Baichuan2-13B-Chat & Baichuan2-7B-Base & Baichuan\\ \hline
\end{tabular}
\end{table}

\begin{figure}[h!]
    \centering
    \includegraphics[width=0.45\textwidth]{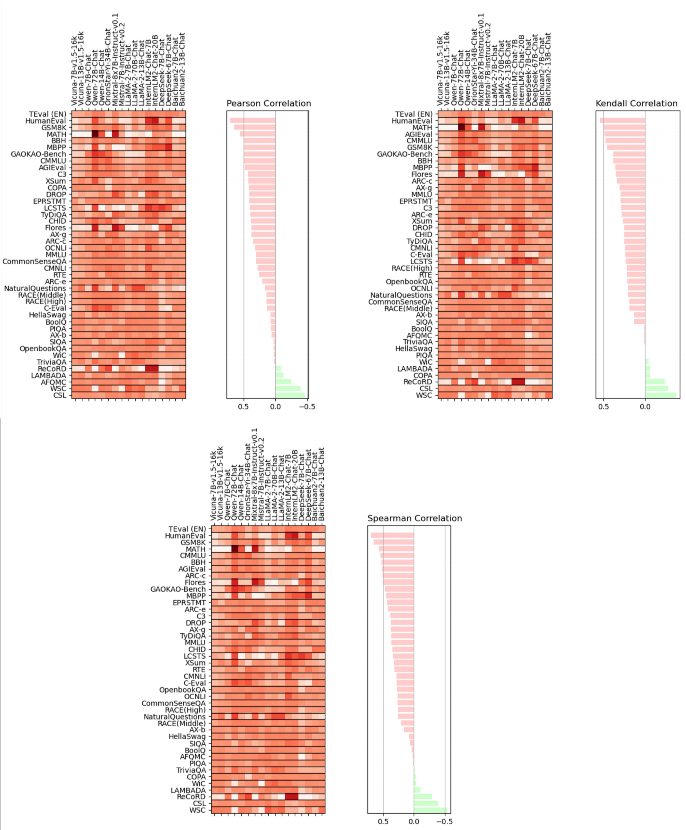}
    \caption{Task relevance measured using Pearson, Spearman, and Kendall correlation metrics.}
    \label{fig:featurevec}
\end{figure}

The task relevance was measured with the Pearson correlation coefficient, the Spearman rank correlation coefficient, and the Kendall rank correlation coefficient respectively. The experimental results are shown in the Figure~\ref{fig:featurevec}.

The tasks most closely related to the T-eval task are primarily reasoning and problem-solving tasks. According to the Pearson correlation coefficient, 5 reasoning tasks, 3 problem-solving tasks, and 2 comprehension tasks rank among the top 10 most similar tasks to T-eval. Under Kendall and Spearman correlation measures, the top 10 include 5 reasoning tasks, 4 problem-solving tasks, and 1 language task. These results suggest that reasoning and problem-solving abilities are strongly correlated with the T-eval task, which aligns with the nature and structure of tool-use tasks.

The comparison of the three correlation metrics, as presented in Table~\ref{tab:corr}, indicates that Kendall's rank correlation consistently provides superior ranking stability across both baseline and sampling consistency experiments. Kendall's coefficient outperforms the other methods, particularly when the number of sampled models and iterations is relatively small, and continues to lead as these parameters increase.

\begin{table}[htbp]
\centering
\footnotesize
\caption{Baseline and Sampling Consistency Experimental Results (P: Pearson correlation, SR: Spearman Rank correlation, KR: Kendall Rank correlation, $t=10$)}
\label{tab:corr}
\begin{tabular}{ccccc}
\toprule
\textbf{Configurations} & \textbf{P} & \textbf{SR} & \textbf{KR} \\
\midrule
$n=6, k=25$ & 0.444 / 0.359 & 0.444 / 0.325 & \textbf{0.492} / \textbf{0.372} \\
$n=8, k=25$ & 0.544 / 0.418 & 0.516 / 0.392 & \textbf{0.548} / \textbf{0.431} \\
$n=10, k=25$ & 0.500 / \textbf{0.476} & 0.568 / 0.472 & \textbf{0.580} / 0.475 \\
$n=10, k=15$ & 0.560 / 0.467 & 0.613 / 0.491 & \textbf{0.640} / \textbf{0.522} \\
$n=10, k=35$ & 0.551 / 0.435 & 0.574 / 0.434 & \textbf{0.574} / \textbf{0.457} \\
\bottomrule
\end{tabular}
\end{table}

Kendall's rank correlation, which measures the monotonic relationship between variables, is more robust in reflecting the inherent relationships between language tasks. This is particularly relevant given that a model's performance on one task often correlates with its performance on related tasks, though not necessarily in a linear fashion.

Compared to the Pearson correlation, Kendall's rank correlation adapts better to the diverse characteristics of language tasks. Moreover, its reduced sensitivity to data distribution further supports its effectiveness in providing a more equitable comparison of task relevance, especially when tasks vary in difficulty and the distribution of model scores differs across tasks.

\subsection{Results for Robustness-Based Selection}
In this robustness experiment, we employed two groups of small models—namely the Data Variability Group, comprising 5 models, and the Random Noise Group, comprising 3 models—to validate task robustness. Both groups were based on the Qwen-1.8B architecture~\cite{bai2023qwen}, and were configured with identical structural and training settings. Specifically, the models utilized Byte Pair Encoding (BPE)\cite{shibata1999byte} for tokenization, Rotary Positional Embedding (RoPE)\cite{su2024roformer} for positional encoding, RMSNorm\cite{zhang2019root} as the pre-normalization technique, and SwiGLU\cite{shazeer2020glu} units as the activation function. For the Random Noise Group, three small models (labeled A, B, and C) were pre-trained on a 30B token subset of the Falcon RefineWeb dataset\cite{penedo2023refinedweb}, with random noise introduced by shuffling this subset before training. The Data Variability Group included five small models (labeled D, E, F, G, and H), each trained on a distinct 100B-scale subset from five publicly available pre-training datasets: Falcon RefineWeb\cite{penedo2023refinedweb}, RedPajamaV2\cite{okazaki2024building}, Wikipedia\cite{merity2016pointer}, BooksCorpus\cite{zhu2015aligning}, and CodeParrot\cite{xu2022systematic}.

After pre-training the eight small models across both groups, each model was evaluated on six tasks, and a robustness score was calculated for each task. The evaluation tasks included C3\cite{sun2020investigating}, CMNLI\cite{xu2020clue}, OCNLI\cite{hu2020ocnli}, CHID\cite{zheng2019chid}, RTE\cite{dagan2010recognizing}, and CMMLU\cite{li2023cmmlu}. These tasks, relevant to the T-eval task in the task relevance evaluation, encompass natural language inference, knowledge-based question answering, and reading comprehension, and adhere to a multiple-choice evaluation paradigm suited for early models. 
\begin{table}[htbp]
\centering
\scriptsize
\caption{Performances of Small Models on Candidate Tasks}
\label{tab:per6}
\begin{tabular}{ccccccccc}
\toprule
\textbf{Task} & \multicolumn{3}{c}{\textbf{Random Noise Group}} & \multicolumn{5}{c}{\textbf{Data Variability Group}} \\
\cmidrule(lr){2-4} \cmidrule(lr){5-9}
 & \textbf{A} & \textbf{B} & \textbf{C} & \textbf{D} & \textbf{E} & \textbf{F} & \textbf{G} & \textbf{H} \\
\midrule
C3 & 31.18 & 30.36 & 32.26 & 35.73 & 46.47 & 37.97 & 26.40 & 40.93 \\
CMNLI & 31.78 & 32.99 & 32.05 & 41.00 & 45.32 & 34.32 & 40.27 & 32.57 \\
OCNLI & 21.48 & 23.43 & 23.63 & 30.00 & 36.05 & 30.18 & 46.05 & 33.30 \\
CHID & 43.83 & 44.68 & 46.52 & 52.20 & 71.98 & 50.15 & 51.46 & 48.01 \\
RTE & 48.38 & 45.85 & 46.36 & 46.38 & 53.50 & 46.86 & 51.56 & 46.01 \\
CMMLU & 25.20 & 24.82 & 25.05 & 24.75 & 26.13 & 25.31 & 25.44 & 24.53 \\
\bottomrule
\end{tabular}
\label{tab:small_model_group_eval}
\end{table}

\begin{table}[htbp]
\centering
\footnotesize
\caption{Task Robustness Analysis Results. $\sigma_r^2$ and $\sigma_d^2$ denotes the variance of the Random Noise Group and Data Variability Group respectively, and $R_i$ denotes the robustness score.}
\label{tab:rob}
\begin{tabular}{cccc}
\toprule
\textbf{Task} & \textbf{$\sigma_r^2$} & \textbf{$\sigma_d^2$} & \textbf{$R_i$} \\
\midrule
C3 & 0.91 & 51.97 & 57.23 \\
CMNLI & 0.40 & 1.18 & 2.94 \\
OCNLI & 8.23 & 7.20 & 0.88 \\
CHID & 11.27 & 577.55 & 51.43 \\
RTE & 1.71 & 8.37 & 4.90 \\
CMMLU & 0.04 & 3.49 & 10.79 \\
\bottomrule
\end{tabular}
\label{tab:task_robustness_analysis}
\end{table}

The detailed performances and robustness analysis results of small models are summarized in Table~\ref{tab:per6} and \ref{tab:rob}.
The statistics revealed that the reading comprehension tasks, C3 \cite{sun2020investigating} and CHID \cite{zheng2019chid}, achieved the highest robustness scores, indicating superior noise resistance at the small model stage. These tasks serve as crucial benchmarks in this context. In contrast, the question-answering task CMMLU \cite{li2023cmmlu}, due to its inherent difficulty, resulted in lower absolute evaluation scores across all small models. Among the language inference tasks, RTE \cite{dagan2010recognizing} exhibited the best noise resistance. These robustness scores offer valuable insights for analyzing the relative importance of each proxy task.

\subsection{Results for Tool-use Capability Prediction}
To validate the effectiveness of prediction approach, different experimental setups were used to train models from scratch. Early in training, models were assessed using proxy task metrics, while later stages involved direct evaluation using the T-eval task. The consistency between these evaluation results was then compared. Kendall's rank correlation coefficient was used to measure task relevance.

We firstly compare the impact of learning rate annealing on model performance. Two configurations were tested:
\begin{itemize}
    \item \textbf{Non-Annealing Group}: An 18-billion-parameter Qwen model pretrained on 3.6 trillion tokens with a Cosine Learning Rate Scheduler applied throughout.
    \item \textbf{Annealing Group}: The same model, but with an annealing operation applied from 2.6 trillion to 3.6 trillion tokens, gradually reducing the learning rate to zero.
\end{itemize}

Model checkpoints from early pretraining stages were evaluated using the proxy task set. After completed pretraining, the models underwent Supervised Fine-Tuning (SFT) and were directly evaluated on the T-eval task. Results in Table~\ref{tab:proana} and \ref{tab:tana} showed that the Annealing Group outperformed the Non-Annealing Group in both proxy task metrics and direct evaluations after SFT. This consistency confirms the reliability of the proxy task metrics and highlights the effectiveness of the annealing strategy in enhancing downstream task performance.

\begin{table}[htbp]
\centering
\footnotesize
\caption{Model Performances on Proxy Tasks under Non-Annealing (NA) and Annealing (A) Group after training $t_1$ and $t_2$ tokens}
\label{tab:proana}
\begin{tabular}{lcccc}
\toprule
\textbf{Test Set} & \textbf{A-$t_1$} & \textbf{NA-$t_1$} & \textbf{A-$t_2$} & \textbf{NA-$t_2$} \\
\midrule
C3 & 51.34 & 50.68 & 50.58 & 49.04 \\
CMNLI & 32.71 & 33.07 & 32.53 & 32.57 \\
OCNLI & 36.57 & 36.47 & 36.80 & 36.80 \\
CHID & 80.82 & 78.52 & 81.62 & 79.47 \\
RTE & 49.46 & 52.35 & 49.10 & 53.43 \\
CMMLU & 25.46 & 25.44 & 25.13 & 25.43 \\
\midrule
\textbf{Proxy Task (↑)} & \textbf{48.03} & 47.77 & \textbf{47.91} & 47.69 \\
\bottomrule
\end{tabular}
\end{table}

\begin{table}[h]
\centering
\footnotesize
\caption{Evaluation Results of Fine-tuned Models of Non-Annealing (NA) and Annealing (A) Group on T-eval}
\label{tab:tana}
\begin{tabular}{c|c|c}
\hline
\textbf{Group} & \textbf{A-3600B-SFT} & \textbf{NA-3600B-SFT} \\ \hline
T-eval Results (↑) & 23.86 & \textbf{22.76}  \\ \hline
\end{tabular}
\end{table}


We also compare the impact of different pretraining data mixtures through three data selection strategies:

\begin{itemize}
    \item \textbf{PPL Group}: This group focused on selecting data with the lowest perplexity from public pretraining corpora, primarily using the Redpajama dataset~\cite{okazaki2024building}, supplemented with Wikipedia, code, and other data types.
    \item \textbf{Filtered Group}: This group employed strict data quality filtering processes.
    \item \textbf{Diversity Group}: Building on the Filtered Group, this approach ensured high-quality filtering while incorporating diverse data sources to enhance variability.
\end{itemize}

Model checkpoints were evaluated at 500 billion tokens of pretraining using the proxy task set, followed by supervised fine-tuning and direct evaluation on the T-eval task.

Results revealed that during pretraining, the Diversity Group achieved the highest scores on proxy task metrics, followed by the Filtered Group, with the PPL Group performing the worst. In the fine-tuned model phase, direct T-eval results ranked the Filtered Group highest, followed by the Diversity Group, while the PPL Group consistently underperformed.

These findings suggest that relying solely on perplexity-based filtering is inadequate for optimizing pretraining data. High-quality filtering and data diversity both positively impact dialogue model performance in tool utilization tasks. Additionally, there was significant consistency between proxy task metrics and direct T-eval results, reinforcing the validity of the proxy evaluation approach.

\begin{table}[h]
\centering
\footnotesize
\caption{Model Performances on Proxy Tasks under PPL, Filtered and Diversity Group after training 500B tokens}
\begin{tabular}{c|c|c|c}
\hline
\textbf{Test Set} & \textbf{PPL} & \textbf{Filter} & \textbf{Diversity} \\ \hline
C3 & 47.62 & 48.33 & 46.19 \\ 
CMNLI & 32.98 & 33.00 & 33.59 \\ 
OCNLI & 29.97 & 30.03 & 35.13 \\ 
CHID & 74.48 & 74.68 & 75.52 \\ 
RTE & 47.65 & 48.01 & 49.46 \\ 
CMMLU & 24.73 & 25.35 & 24.71 \\ \hline
\textbf{Proxy Task} (↑) & 45.01 & 45.40 & \textbf{45.57} \\ \hline
\end{tabular}
\end{table}

\begin{table}[h]
\centering
\footnotesize
\caption{Evaluation Results of Fine-tuned Models of PPL, Filtered and Diversity Group on T-eval}
\begin{tabular}{c|c|c|c}
\hline
\textbf{Model} & \textbf{PPL} & \textbf{Filter} & \textbf{Diversity} \\ \hline
T-eval Results (↑) & 15.47 & \textbf{17.43} & 17.02 \\ \hline
\end{tabular}
\end{table}

We compared three strategies for selecting proxy tasks and predicting tool-use performance. The first strategy, $V_{chat}$, uses chat models for both selection and prediction. The second, $V_{base}$, relies solely on base models. The third, $V_{bc}$, combines both: chat models for dialogue tasks and base models for prediction.

\begin{table}[ht]
\centering
\scriptsize
\caption{Prediction results with three strategies to use proxy tasks or training PPL, detailing \textbf{score/rank} for the 5 pretrained models and num of reverse order pairs (\textbf{RO}) with ground truth (T-eval ranking).}
\label{tab:data_ratios}
\begin{tabular}{c|c|c|c|c|c|c}
\hline
         & \textbf{PPL}  & \textbf{Filtered}  & \textbf{Diversity}  & \textbf{A} & \textbf{NA} & \textbf{RO} (↓) \\ \hline
T-eval (↑)    & 15.47/5   & 17.43/3        & 17.02/4         & 23.86/1        & 22.76/2            &  -            \\ \hline
PPL (↓)       & 3.55/2    & 3.98/4         & 4.03/5          & 3.57/3         & 3.48/1             & 4/10              \\ \hline
$V_{chat}$  (↑) & 38.43/5   & 38.87/4        & 38.93/3         & 40.45/2        & 40.69/1            & 2/10              \\ 
$V_{base}$  (↑) & 28.65/3   & 29.46/2        & 27.72/5         & 28.00/4        & 30.03/1            & 4/10              \\ 
$V_{bc}$ (↑)   & 45.01/5   & 45.40/4        & 45.57/3         & 47.91/1        & 47.69/2            & \textbf{1/10}              \\ \hline
\end{tabular}
\end{table}

As illustrated in Table~\ref{tab:data_ratios}, the number of inverse order pairs compared to T-eval rankings reflects the consistency of tool-use capability rankings across the five models. Notably, the $V_{bc}$ strategy yields rankings that closely align with those from T-eval. Although PPL is commonly used to evaluate early checkpoints for optimizing training settings, these results underscore the instability and potential unreliability of relying solely on PPL.

%% file: latex/conclusion.tex
\section{Conclusion}

We introduced a method to predict emergent abilities in LLMs by utilizing proxy tasks, addressing the limitations of scaling laws in forecasting capabilities that do not manifest in smaller or early-stage models. By establishing relevance metrics between target tasks and candidate tasks, and validating these tasks for robustness with small model ensembles, we identified the most suitable proxies for early-stage evaluation. The predicted performance on the target task is then derived from these proxies, offering a more reliable forecast of the model's capabilities.

Our approach demonstrated strong predictive power, particularly in the case study on tool utilization, where there was a high correlation between the predicted and actual performance of LLMs. This method not only allows for accurate early-stage predictions of complex abilities but also provides valuable insights for optimizing training configurations, significantly improving the efficiency and reliability of LLM development.

By shifting the evaluation focus to the early stages of training, this method addresses key challenges in predicting emergent abilities, offering a robust framework for enhancing LLM performance in real-world applications.

%% file: acl_latex.bbl
\begin{thebibliography}{38}
\providecommand{\natexlab}[1]{#1}

\bibitem[{Achiam et~al.(2023)Achiam, Adler, Agarwal, Ahmad, Akkaya, Aleman, Almeida, Altenschmidt, Altman, Anadkat et~al.}]{achiam2023gpt}
Josh Achiam, Steven Adler, Sandhini Agarwal, Lama Ahmad, Ilge Akkaya, Florencia~Leoni Aleman, Diogo Almeida, Janko Altenschmidt, Sam Altman, Shyamal Anadkat, et~al. 2023.
\newblock Gpt-4 technical report.
\newblock \emph{arXiv preprint arXiv:2303.08774}.

\bibitem[{Almazrouei et~al.(2023)Almazrouei, Alobeidli, Alshamsi, Cappelli, Cojocaru, Debbah, Goffinet, Hesslow, Launay, Malartic et~al.}]{almazrouei2023falcon}
Ebtesam Almazrouei, Hamza Alobeidli, Abdulaziz Alshamsi, Alessandro Cappelli, Ruxandra Cojocaru, M{\'e}rouane Debbah, {\'E}tienne Goffinet, Daniel Hesslow, Julien Launay, Quentin Malartic, et~al. 2023.
\newblock The falcon series of open language models.
\newblock \emph{arXiv preprint arXiv:2311.16867}.

\bibitem[{Anwar et~al.(2024)Anwar, Saparov, Rando, Paleka, Turpin, Hase, Lubana, Jenner, Casper, Sourbut et~al.}]{anwar2024foundational}
Usman Anwar, Abulhair Saparov, Javier Rando, Daniel Paleka, Miles Turpin, Peter Hase, Ekdeep~Singh Lubana, Erik Jenner, Stephen Casper, Oliver Sourbut, et~al. 2024.
\newblock Foundational challenges in assuring alignment and safety of large language models.
\newblock \emph{arXiv preprint arXiv:2404.09932}.

\bibitem[{Arora and Goyal(2023)}]{arora2023theory}
Sanjeev Arora and Anirudh Goyal. 2023.
\newblock A theory for emergence of complex skills in language models.
\newblock \emph{arXiv preprint arXiv:2307.15936}.

\bibitem[{Bai et~al.(2023)Bai, Bai, Chu, Cui, Dang, Deng, Fan, Ge, Han, Huang et~al.}]{bai2023qwen}
Jinze Bai, Shuai Bai, Yunfei Chu, Zeyu Cui, Kai Dang, Xiaodong Deng, Yang Fan, Wenbin Ge, Yu~Han, Fei Huang, et~al. 2023.
\newblock Qwen technical report.
\newblock \emph{arXiv preprint arXiv:2309.16609}.

\bibitem[{Brown et~al.(2020)Brown, Mann, Ryder, Subbiah, Kaplan, Dhariwal, Neelakantan, Shyam, Sastry, Askell et~al.}]{brown2020language}
Tom Brown, Benjamin Mann, Nick Ryder, Melanie Subbiah, Jared~D Kaplan, Prafulla Dhariwal, Arvind Neelakantan, Pranav Shyam, Girish Sastry, Amanda Askell, et~al. 2020.
\newblock Language models are few-shot learners.
\newblock \emph{Advances in neural information processing systems}, 33:1877--1901.

\bibitem[{Buitrago and Nystrom(2019)}]{buitrago2019open}
Paola~A Buitrago and Nicholas~A Nystrom. 2019.
\newblock Open compass: accelerating the adoption of ai in open research.
\newblock pages 1--9.

\bibitem[{Chen et~al.(2023)Chen, Du, Zhang, Liu, Liu, Zheng, Zhuo, Zhang, Lin, Chen et~al.}]{chen2023t}
Zehui Chen, Weihua Du, Wenwei Zhang, Kuikun Liu, Jiangning Liu, Miao Zheng, Jingming Zhuo, Songyang Zhang, Dahua Lin, Kai Chen, et~al. 2023.
\newblock T-eval: Evaluating the tool utilization capability step by step.
\newblock \emph{arXiv preprint arXiv:2312.14033}.

\bibitem[{Dagan et~al.(2010)Dagan, Dolan, Magnini, and Roth}]{dagan2010recognizing}
Ido Dagan, Bill Dolan, Bernardo Magnini, and Dan Roth. 2010.
\newblock Recognizing textual entailment: Rational, evaluation and approaches--erratum.
\newblock \emph{Natural Language Engineering}, 16(1):105--105.

\bibitem[{Du et~al.(2024)Du, Zeng, Dong, and Tang}]{du2024understanding}
Zhengxiao Du, Aohan Zeng, Yuxiao Dong, and Jie Tang. 2024.
\newblock Understanding emergent abilities of language models from the loss perspective.
\newblock \emph{arXiv preprint arXiv:2403.15796}.

\bibitem[{Gadre et~al.(2024)Gadre, Smyrnis, Shankar, Gururangan, Wortsman, Shao, Mercat, Fang, Li, Keh et~al.}]{gadre2024language}
Samir~Yitzhak Gadre, Georgios Smyrnis, Vaishaal Shankar, Suchin Gururangan, Mitchell Wortsman, Rulin Shao, Jean Mercat, Alex Fang, Jeffrey Li, Sedrick Keh, et~al. 2024.
\newblock Language models scale reliably with over-training and on downstream tasks.
\newblock \emph{arXiv preprint arXiv:2403.08540}.

\bibitem[{Ganguli et~al.(2022)Ganguli, Hernandez, Lovitt, Askell, Bai, Chen, Conerly, Dassarma, Drain, Elhage et~al.}]{ganguli2022predictability}
Deep Ganguli, Danny Hernandez, Liane Lovitt, Amanda Askell, Yuntao Bai, Anna Chen, Tom Conerly, Nova Dassarma, Dawn Drain, Nelson Elhage, et~al. 2022.
\newblock Predictability and surprise in large generative models.
\newblock In \emph{Proceedings of the 2022 ACM Conference on Fairness, Accountability, and Transparency}, pages 1747--1764.

\bibitem[{Henighan et~al.(2020)Henighan, Kaplan, Katz, Chen, Hesse, Jackson, Jun, Brown, Dhariwal, Gray et~al.}]{henighan2020scaling}
Tom Henighan, Jared Kaplan, Mor Katz, Mark Chen, Christopher Hesse, Jacob Jackson, Heewoo Jun, Tom~B Brown, Prafulla Dhariwal, Scott Gray, et~al. 2020.
\newblock Scaling laws for autoregressive generative modeling.
\newblock \emph{arXiv preprint arXiv:2010.14701}.

\bibitem[{Hu et~al.(2020)Hu, Richardson, Xu, Li, K{\"u}bler, and Moss}]{hu2020ocnli}
Hai Hu, Kyle Richardson, Liang Xu, Lu~Li, Sandra K{\"u}bler, and Lawrence~S Moss. 2020.
\newblock Ocnli: Original chinese natural language inference.
\newblock \emph{arXiv preprint arXiv:2010.05444}.

\bibitem[{Hu et~al.(2023)Hu, Liu, Han, Zhang, He, Zhao, Lin, Ding, Ou, Zeng et~al.}]{hu2023predicting}
Shengding Hu, Xin Liu, Xu~Han, Xinrong Zhang, Chaoqun He, Weilin Zhao, Yankai Lin, Ning Ding, Zebin Ou, Guoyang Zeng, et~al. 2023.
\newblock Predicting emergent abilities with infinite resolution evaluation.
\newblock In \emph{The Twelfth International Conference on Learning Representations}.

\bibitem[{Huang et~al.(2024)Huang, Zhang, Shan, and He}]{huang2024compression}
Yuzhen Huang, Jinghan Zhang, Zifei Shan, and Junxian He. 2024.
\newblock Compression represents intelligence linearly.
\newblock \emph{arXiv preprint arXiv:2404.09937}.

\bibitem[{Li et~al.(2023)Li, Zhang, Koto, Yang, Zhao, Gong, Duan, and Baldwin}]{li2023cmmlu}
Haonan Li, Yixuan Zhang, Fajri Koto, Yifei Yang, Hai Zhao, Yeyun Gong, Nan Duan, and Timothy Baldwin. 2023.
\newblock Cmmlu: Measuring massive multitask language understanding in chinese.
\newblock \emph{arXiv preprint arXiv:2306.09212}.

\bibitem[{Ling et~al.(2023)Ling, Wu, Dong, Feng, Karypis, and Reddy}]{ling2023international}
Yuan Ling, Fanyou Wu, Shujing Dong, Yarong Feng, George Karypis, and Chandan~K Reddy. 2023.
\newblock International workshop on multimodal learning-2023 theme: Multimodal learning with foundation models.
\newblock In \emph{Proceedings of the 29th ACM SIGKDD Conference on Knowledge Discovery and Data Mining}, pages 5868--5869.

\bibitem[{Lu et~al.(2023)Lu, Bigoulaeva, Sachdeva, Madabushi, and Gurevych}]{lu2023emergent}
Sheng Lu, Irina Bigoulaeva, Rachneet Sachdeva, Harish~Tayyar Madabushi, and Iryna Gurevych. 2023.
\newblock Are emergent abilities in large language models just in-context learning?
\newblock \emph{arXiv preprint arXiv:2309.01809}.

\bibitem[{Merity et~al.(2016)Merity, Xiong, Bradbury, and Socher}]{merity2016pointer}
Stephen Merity, Caiming Xiong, James Bradbury, and Richard Socher. 2016.
\newblock Pointer sentinel mixture models.
\newblock \emph{arXiv preprint arXiv:1609.07843}.

\bibitem[{Okazaki et~al.(2024)Okazaki, Hattori, Shota, Iida, Ohi, Fujii, Nakamura, Loem, Yokota, and Mizuki}]{okazaki2024building}
Naoaki Okazaki, Kakeru Hattori, Hirai Shota, Hiroki Iida, Masanari Ohi, Kazuki Fujii, Taishi Nakamura, Mengsay Loem, Rio Yokota, and Sakae Mizuki. 2024.
\newblock Building a large japanese web corpus for large language models.
\newblock \emph{arXiv preprint arXiv:2404.17733}.

\bibitem[{Penedo et~al.(2023)Penedo, Malartic, Hesslow, Cojocaru, Alobeidli, Cappelli, Pannier, Almazrouei, and Launay}]{penedo2023refinedweb}
Guilherme Penedo, Quentin Malartic, Daniel Hesslow, Ruxandra Cojocaru, Hamza Alobeidli, Alessandro Cappelli, Baptiste Pannier, Ebtesam Almazrouei, and Julien Launay. 2023.
\newblock The refinedweb dataset for falcon llm: Outperforming curated corpora with web data only.
\newblock \emph{Advances in Neural Information Processing Systems}, 36:79155--79172.

\bibitem[{Shannon(1948)}]{shannon1948mathematical}
Claude~Elwood Shannon. 1948.
\newblock A mathematical theory of communication.
\newblock \emph{The Bell system technical journal}, 27(3):379--423.

\bibitem[{Shazeer(2020)}]{shazeer2020glu}
Noam Shazeer. 2020.
\newblock Glu variants improve transformer.
\newblock \emph{arXiv preprint arXiv:2002.05202}.

\bibitem[{Shibata et~al.(1999)Shibata, Kida, Fukamachi, Takeda, Shinohara, Shinohara, and Arikawa}]{shibata1999byte}
Yusuxke Shibata, Takuya Kida, Shuichi Fukamachi, Masayuki Takeda, Ayumi Shinohara, Takeshi Shinohara, and Setsuo Arikawa. 1999.
\newblock Byte pair encoding: A text compression scheme that accelerates pattern matching.

\bibitem[{Srivastava et~al.(2022)Srivastava, Rastogi, Rao, Shoeb, Abid, Fisch, Brown, Santoro, Gupta, Garriga-Alonso et~al.}]{srivastava2022beyond}
Aarohi Srivastava, Abhinav Rastogi, Abhishek Rao, Abu Awal~Md Shoeb, Abubakar Abid, Adam Fisch, Adam~R Brown, Adam Santoro, Aditya Gupta, Adri{\`a} Garriga-Alonso, et~al. 2022.
\newblock Beyond the imitation game: Quantifying and extrapolating the capabilities of language models.
\newblock \emph{arXiv preprint arXiv:2206.04615}.

\bibitem[{Su et~al.(2024)Su, Ahmed, Lu, Pan, Bo, and Liu}]{su2024roformer}
Jianlin Su, Murtadha Ahmed, Yu~Lu, Shengfeng Pan, Wen Bo, and Yunfeng Liu. 2024.
\newblock Roformer: Enhanced transformer with rotary position embedding.
\newblock \emph{Neurocomputing}, 568:127063.

\bibitem[{Sun et~al.(2020)Sun, Yu, Yu, and Cardie}]{sun2020investigating}
Kai Sun, Dian Yu, Dong Yu, and Claire Cardie. 2020.
\newblock Investigating prior knowledge for challenging chinese machine reading comprehension.
\newblock \emph{Transactions of the Association for Computational Linguistics}, 8:141--155.

\bibitem[{Suzgun et~al.(2022)Suzgun, Scales, Sch{\"a}rli, Gehrmann, Tay, Chung, Chowdhery, Le, Chi, Zhou et~al.}]{suzgun2022challenging}
Mirac Suzgun, Nathan Scales, Nathanael Sch{\"a}rli, Sebastian Gehrmann, Yi~Tay, Hyung~Won Chung, Aakanksha Chowdhery, Quoc~V Le, Ed~H Chi, Denny Zhou, et~al. 2022.
\newblock Challenging big-bench tasks and whether chain-of-thought can solve them.
\newblock \emph{arXiv preprint arXiv:2210.09261}.

\bibitem[{Wei et~al.(2022{\natexlab{a}})Wei, Tay, Bommasani, Raffel, Zoph, Borgeaud, Yogatama, Bosma, Zhou, Metzler et~al.}]{wei2022emergent}
Jason Wei, Yi~Tay, Rishi Bommasani, Colin Raffel, Barret Zoph, Sebastian Borgeaud, Dani Yogatama, Maarten Bosma, Denny Zhou, Donald Metzler, et~al. 2022{\natexlab{a}}.
\newblock Emergent abilities of large language models.
\newblock \emph{arXiv preprint arXiv:2206.07682}.

\bibitem[{Wei et~al.(2022{\natexlab{b}})Wei, Wang, Schuurmans, Bosma, Xia, Chi, Le, Zhou et~al.}]{wei2022chain}
Jason Wei, Xuezhi Wang, Dale Schuurmans, Maarten Bosma, Fei Xia, Ed~Chi, Quoc~V Le, Denny Zhou, et~al. 2022{\natexlab{b}}.
\newblock Chain-of-thought prompting elicits reasoning in large language models.
\newblock \emph{Advances in neural information processing systems}, 35:24824--24837.

\bibitem[{Xia et~al.(2022)Xia, Artetxe, Zhou, Lin, Pasunuru, Chen, Zettlemoyer, and Stoyanov}]{xia2022training}
Mengzhou Xia, Mikel Artetxe, Chunting Zhou, Xi~Victoria Lin, Ramakanth Pasunuru, Danqi Chen, Luke Zettlemoyer, and Ves Stoyanov. 2022.
\newblock Training trajectories of language models across scales.
\newblock \emph{arXiv preprint arXiv:2212.09803}.

\bibitem[{Xie et~al.(2024)Xie, Pham, Dong, Du, Liu, Lu, Liang, Le, Ma, and Yu}]{xie2024doremi}
Sang~Michael Xie, Hieu Pham, Xuanyi Dong, Nan Du, Hanxiao Liu, Yifeng Lu, Percy~S Liang, Quoc~V Le, Tengyu Ma, and Adams~Wei Yu. 2024.
\newblock Doremi: Optimizing data mixtures speeds up language model pretraining.
\newblock \emph{Advances in Neural Information Processing Systems}, 36.

\bibitem[{Xu et~al.(2022)Xu, Alon, Neubig, and Hellendoorn}]{xu2022systematic}
Frank~F Xu, Uri Alon, Graham Neubig, and Vincent~Josua Hellendoorn. 2022.
\newblock A systematic evaluation of large language models of code.
\newblock In \emph{Proceedings of the 6th ACM SIGPLAN International Symposium on Machine Programming}, pages 1--10.

\bibitem[{Xu et~al.(2020)Xu, Hu, Zhang, Li, Cao, Li, Xu, Sun, Yu, Yu et~al.}]{xu2020clue}
Liang Xu, Hai Hu, Xuanwei Zhang, Lu~Li, Chenjie Cao, Yudong Li, Yechen Xu, Kai Sun, Dian Yu, Cong Yu, et~al. 2020.
\newblock Clue: A chinese language understanding evaluation benchmark.
\newblock \emph{arXiv preprint arXiv:2004.05986}.

\bibitem[{Zhang and Sennrich(2019)}]{zhang2019root}
Biao Zhang and Rico Sennrich. 2019.
\newblock Root mean square layer normalization.
\newblock \emph{Advances in Neural Information Processing Systems}, 32.

\bibitem[{Zheng et~al.(2019)Zheng, Huang, and Sun}]{zheng2019chid}
Chujie Zheng, Minlie Huang, and Aixin Sun. 2019.
\newblock Chid: A large-scale chinese idiom dataset for cloze test.
\newblock \emph{arXiv preprint arXiv:1906.01265}.

\bibitem[{Zhu(2015)}]{zhu2015aligning}
Yukun Zhu. 2015.
\newblock Aligning books and movies: Towards story-like visual explanations by watching movies and reading books.
\newblock \emph{arXiv preprint arXiv:1506.06724}.

\end{thebibliography}
